\documentclass{article}
\usepackage{ijcai23}

\usepackage{times}
\usepackage{soul}
\usepackage{url}
\usepackage[hidelinks]{hyperref}
\usepackage[utf8]{inputenc}
\usepackage[small]{caption}
\usepackage{graphicx}
\usepackage{amsmath}
\usepackage{amsthm}
\usepackage{booktabs}
\usepackage{algorithm}
\usepackage{algorithmic}
\usepackage[switch]{lineno}

\usepackage{bbding}
\usepackage{amssymb}
\usepackage{multirow}
\title{Action Recognition with Multi-stream Motion Modeling and Mutual Information Maximization}

\author{
Yuheng Yang$^1$
\and
Haipeng Chen$^1$\and
Zhenguang Liu $^{2*}$\and
Yingda Lyu$^{3*}$\and
Beibei Zhang$^5$\and
Shuang Wu$^{4*}$\and
Zhibo Wang$^2$\And
Kui Ren $^2$
\affiliations
$^1$College of Computer Science and Technology, Jilin University\\
$^2$School of Cyber Science and Technology, Zhejiang University \\
$^3$Public Computer Education and Research Center, Jilin University\\
$^4$Black Sesame Technologies\\
$^5$Zhejiang Lab \\
\emails
yangyh20@mails.jlu.edu.cn,
\{chenhp, ydlv\}@jlu.edu.cn,
\{liuzhenguang2008, bzeecs\}@gmail.com,
wushuang@outlook.sg,
\{zhibowang, kuiren\}@zju.edu.cn
}

\begin{document}

\maketitle

\begin{abstract}
    Action recognition has long been a fundamental and intriguing problem in artificial intelligence. The task is challenging due to the high dimensionality nature of an action, as well as the subtle motion details to be considered. Current state-of-the-art approaches typically learn from articulated motion sequences in the straightforward 3D Euclidean space. However, the \emph{vanilla} Euclidean space is not efficient for modeling important motion characteristics such as the joint-wise angular acceleration, which reveals the driving force behind the motion. Moreover, current methods typically attend to each channel equally and lack theoretical constrains on extracting task-relevant features from the input. 

\quad\; In this paper, we seek to tackle these challenges from three aspects: (1) We propose to incorporate an acceleration representation, explicitly modeling the higher-order variations in motion. 
(2) We introduce a novel Stream-GCN network equipped with multi-stream components and channel attention, where different representations (\emph{i.e.}, streams) supplement each other towards a more precise action recognition while attention capitalizes on those important channels. (3) We explore feature-level supervision for maximizing the extraction of task-relevant information and formulate this into a mutual information loss. Empirically, our approach sets the new state-of-the-art performance on three benchmark datasets, NTU RGB+D, NTU RGB+D 120, and NW-UCLA. Our code is anonymously released at https://github.com/ActionR-Group/Stream-GCN, hoping to inspire the community.
\end{abstract}
\section{Introduction}

Having a mental representation of what actions other humans are performing is crucial for us to adjust our behaviors and plan our course of actions~\cite{liu2021aggregated}. Similarly, the capacity for machines to model and recognize human actions is very much coveted. Now, action recognition spawns a wide spectrum of applications including {public space surveillance}~\cite{zhang2018egogesture}, {human-robot interaction}~\cite{wang2016temporal}, {violence detection}~\cite{singh2018eye}, and {autonomous driving}~\cite{zhang2018egogesture}. As such, action recognition has attracted extensive attention in the past decade~\cite{zheng2018unsupervised,shi2019two}.

Fundamentally, action recognition amounts to learning a mapping from a pose sequence to an action class label. The task is inherently challenging due to the high dimensionality of the input pose sequence and the subtle motion details which differentiate different actions.  In this paper, we focalize action recognition with 3D articulated pose sequences as inputs, given that  3D articulated pose can now be conveniently acquired through commodity motion capture systems or extracted from videos via off-the-shelf pose estimation methods like  ~\cite{shotton2012efficient,wang2014cross}. 

\begin{figure}[t]
  \centering
  \includegraphics[scale=0.4]{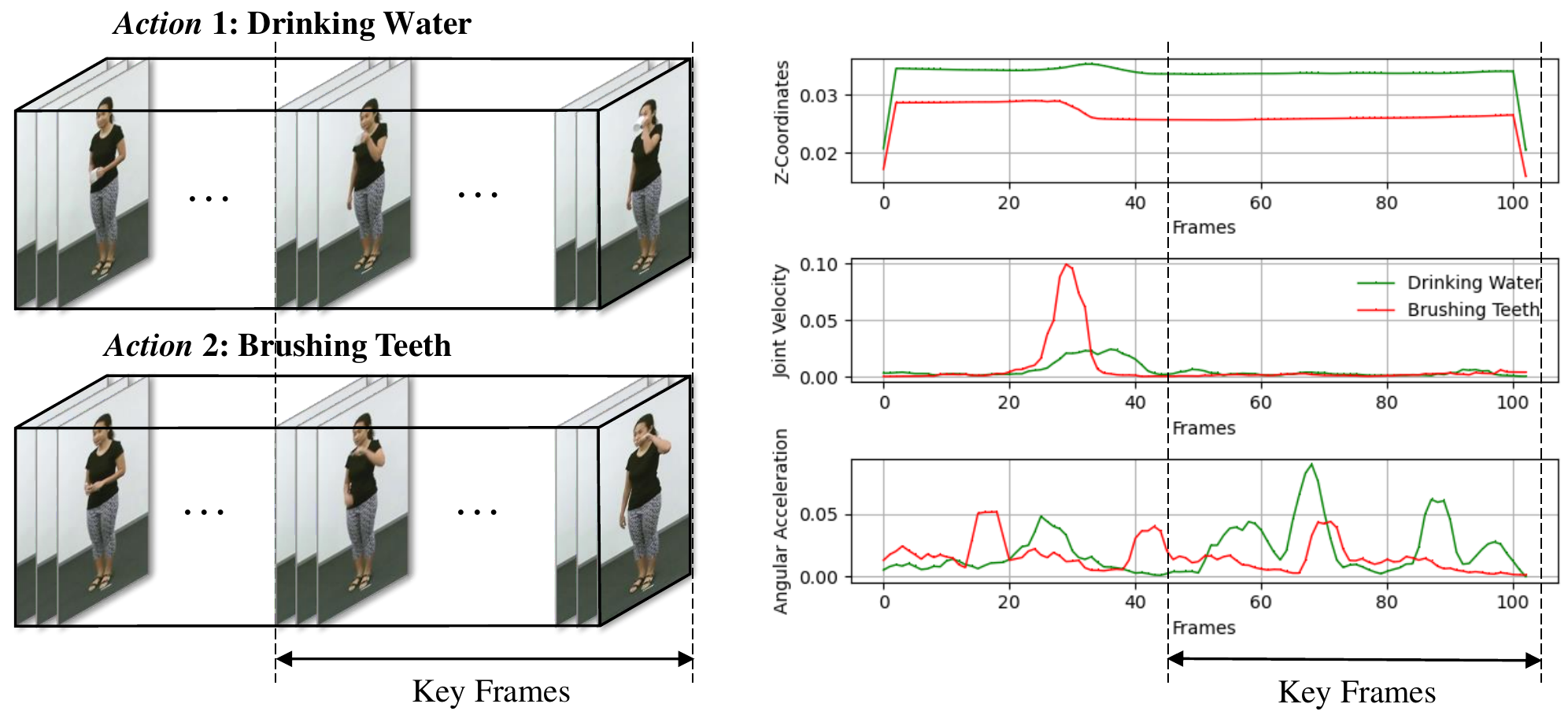}
  \caption{The illustration of similar actions and their features at different orders. \textbf{Left}: Showcase two videos of similar actions, \emph{``drinking water"} and \emph{``brushing teeth"}, the key frames are video clips that are decisive for identifying the action. \textbf{Right}: Take the hand joint as an example, the curves indicate trajectories of motion characteristics for various orders on $z$-dimension.}
  \label{fig1}
\end{figure}

Earlier approaches for action recognition~\cite{li2018independently,vemulapalli2014human} cast the human body as a set of 3D joint coordinates, which in fact treat different joints as independent entities and fail to capture connections between joints~\cite{liu2019towards}. Recently, GCN (Graph Convolutional Network) based methods have become popular solutions as the human skeleton inherently possesses a tree-like graph structure. \textbf{A first class of work} puts the efforts in selecting and enriching features to better represent motion~\cite{shi2019skeleton,shi2019two}, facilitating action recognition from the input feature perspective. For instance, \cite{shi2019two} advocates incorporating the lengths and directions of bones as  second-order input features, which are informative for action recognition. \cite{shi2019skeleton} adds frame-wise bone and joint velocities as additional motion features to enhance the body representation. \textbf{Another line of work} resorts to devising sophisticated adjacency matrices, seeking to pursue better modeling of joint dependencies. For example, \cite{liu2020disentangling} introduces a multi-scale aggregation scheme, leveraging distance adjacency matrices to measure the degree of dependency between two arbitrary skeleton joints. \cite{li2019actional} presents a method to adaptively learn the non-physical dependencies among joints by an encoder-decoder structure, which automatically infers the link strengths between joints.

Upon investigating and experimenting on the released code of state-of-the-art methods~\cite{chen2021channel,yan2018spatial,chen2021multi}, we empirically observe that: \textbf{First}, current approaches typically revolve around employing lower-order features such as joint coordinates or velocities in the  Euclidean space. Despite their simplicity, these features might not be favourable for describing subtle motion details such as joint angular acceleration and motion trends. Figure~\ref{fig1} illustrates two similar actions, ``drinking water" and ``brushing teeth". It is evident to see that the joint coordinates of the two actions are very close to each other, while the velocity features of the two actions are also very similar. However, the joint angular accelerations of the two actions are quite different.  
\textbf{Second}, most existing approaches attend to each channel \emph{equally}, dissevering the fact that {different channels contribute unequally in recognizing an action}. For instance, in identifying the action ``drinking water", vertical movements ($z$-dimension) of the hand joint play a more vital role than horizontal movements ($xy$-dimensions). Apparently, it is not a good idea to assign equal weight to each dimension. \textbf{Third}, current approaches typically employ the conventional CE (Cross-Entropy) loss to supervise the learning of motion information, lacking an effective constraint on guaranteeing that task-relevant features are extracted.

In this paper, we embrace three key designs to tackle the challenges. Technically, \textbf{(1)} We consider incorporating angular accelerations as complementary information and extract them within the framework of rigid body kinematics. These higher-order acceleration features provide an additional input stream, on top of the original Euclidean space motion features, potentially improving the completeness and 
robustness of the model.
\textbf{(2)} We introduce a novel GCN network equipped with multiple input streams and channel-wise attention, where different input streams are assembled while  attention weights capitalize on those important channels. \textbf{(3)} We further engage an information-theoretic objective between the extracted deep features and action labels. Maximizing this mutual information objective drives our model to fully mine task-relevant information while reducing task-irrelevant nuisances.

Thereafter, we conduct extensive experiments on three large benchmark datasets including NTU RGB+D, NTU RGB+D 120, and NW-UCLA. Empirically, our approach consistently and significantly outperforms state-of-the-art methods. Interestingly, we observe that existing methods indeed have difficulties in distinguishing between similar actions. 

To summarize, our \textbf{key contributions} are as follows: 1) A new motion representation is proposed,
which enriches the lower-order motion representations with higher-order motion representations. 2) A multi-stream graph convolution network is presented to comprehensively extract knowledge from multiple representations. Within the framework, the distinct weights of different channels are considered.  To explicitly supervise the knowledge extraction from the input motion sequence, we theoretically analyzed the mutual information between
the extracted deep feature and the label, arriving at a loss that maximizes the task-relevant information. 3) Our method achieves the new state-of-the-art performance on three benchmark datasets and overall provides interesting insights. The implementations are released, hoping to facilitate future research.

\section{Method}

Broadly, a human skeleton can be cast as a set of joints and bones.  We refer to the set of joints as a set of nodes $\mathcal{V}=
\{v_i\}_{i=1}^n$. Further, we model the set of bones, each of which connects a pair of joints, as a set of edges $\mathcal{E}=\{e_i\}_{i=1}^{n \times n}$.
Thus, we conveniently depict a human skeleton pose as a graph $\mathcal{G}=(\mathcal{V}, \mathcal{E})$,
where $n=|\mathcal{V}|$ and $\mathcal{E}\subset \mathcal{V}\times \mathcal{V}$.

\textbf{Problem.} \quad Presented with an observed  3D skeleton motion sequence $\mathcal{S} =\langle \mathcal{G}_1, \mathcal{G}_2, \cdots, \mathcal{G}_m \rangle$, where $\mathcal{G}_i$  is the skeleton pose of the human in the $i^{th}$ frame,
we are interested in predicting its action class label $y \in \mathbb{R}$. Put differently, we seek to learn the non-linear mapping from a motion sequence $\mathcal{S}$  to its action class label $y$.

\textbf{Method Overview.} \quad 
Overall, our method consists of three key components: {a)} motion representations engaging  both lower-order and higher-order motion features, {b)}  a novel Stream-GCN network that is equipped with channel-wise attention, and is able to integrate higher-order and lower-order representations, and {c)} a mutual information objective that effectuates feature-level supervision for maximizing the extraction of task-relevant information. Specifically, we {first} leverage multiple motion representations to frame the motion context. Then, the motion sequences are fed into the proposed Stream-GCN network to predict the action label. Within the network, we introduce a mutual information objective to supervise the prediction. We would like to highlight that our approach has an edge in engaging acceleration to capture subtle motion details, while our network games in maximizing the extraction of task-relevant information and getting rid of task-irrelevant nuisances. In what follows, we elaborate on the three key components in detail.

\subsection{Motion Representation}
\label{2.1}

To represent human motion, there have been efforts to outline a pose as 3D coordinates of all skeletal joints~\cite{liu2019towards}. The 3D joint coordinates  offer an intuitive and direct way to precisely sketch the motion. Unfortunately, such a  representation regards different joints as independent entities and is unfavourable for describing higher-order subtle motion details such as {bone rotation trends}~\cite{shi2019skeleton}.

Motivated by this, we incorporate both higher-order and lower-order motion representations. For lower-order representations, we include joint coordinates, bone lengths, joint velocity, and bone velocity. For higher-order representations, we engage in joint and bone angular acceleration representations. These different representations (\emph{i.e.}, streams) supplement each other towards a more complete and precise motion characterization.  Below, we formulate the joint (or bone) angular acceleration representation.

The articulate human skeleton can be considered as rigid bodies (\emph{i.e.} bones) connected by joints. The motion of a joint in relation to its neighbouring joints can therefore be characterized in the framework of rigid body kinematics. 
Within this framework, the angular acceleration of a joint has a significant role.
Since the  acceleration of an object depends directly on the force acting on it, the acceleration change actually characterizes the changes of forces exerted upon joints, which serve as the driven source for motion.
Take drinking water as an example, conventional methods merely consider the position displacement of the raised arm.
They omit the acceleration change of the arm in the process of lifting the hand, moving upwards with a constant speed, and
decelerating when approaching the mouth. 

\begin{figure}[t]
  \centering
  \includegraphics[scale=0.26]{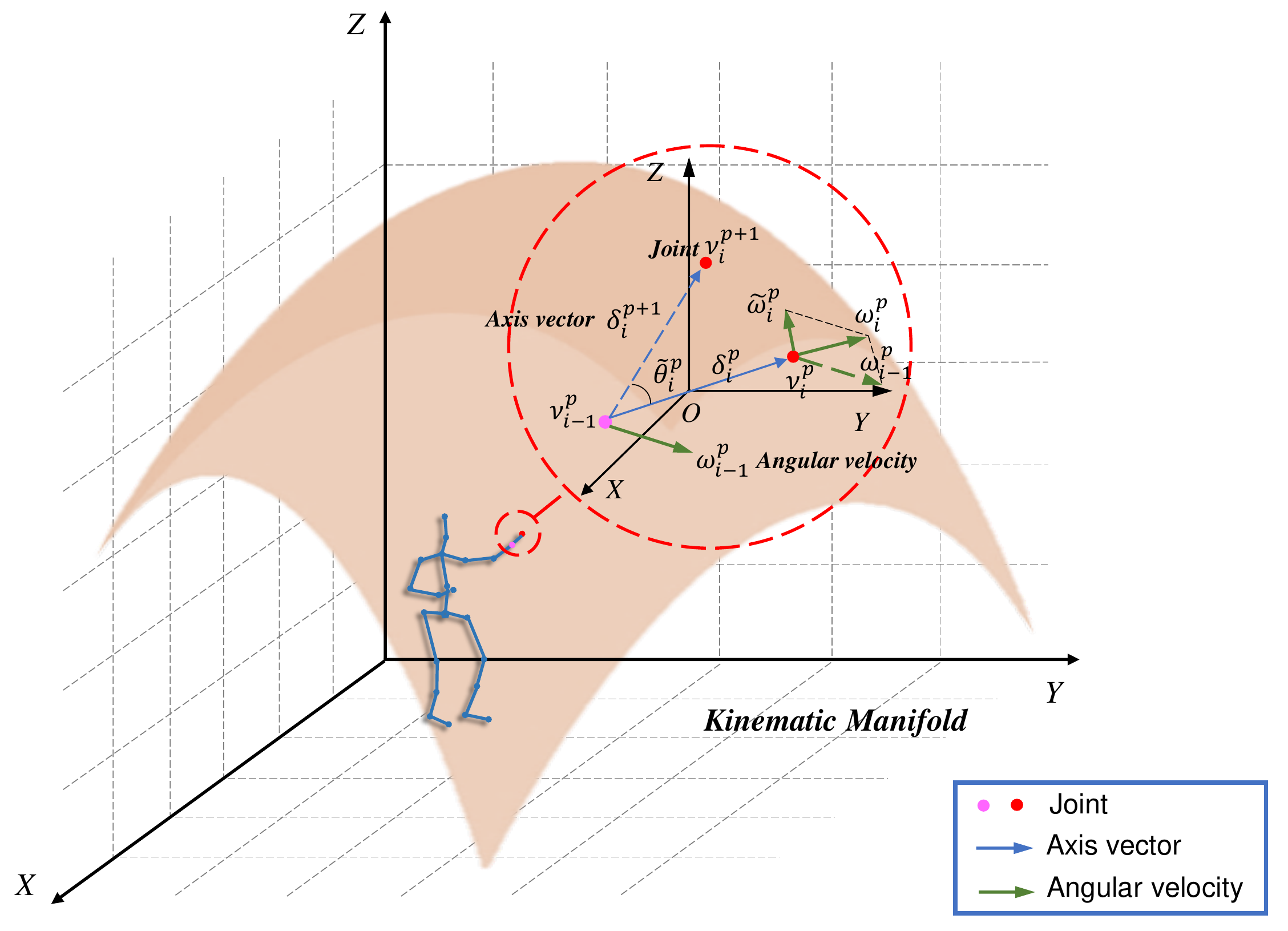}
  \caption{The visualization of \emph{instantaneous angle} $\theta_i^p$ and \emph{instantaneous angular velocity} $\omega_{i}^p$.}\vspace{-1.1em}
  \label{fig2}
\end{figure}

\textbf{Instantaneous Angle.} \quad  We denote   $v_i^p$ as the coordinate of the $i$-th joint at the $p$-th frame. Supposing we are to calculate the acceleration feature of $v_i^p$. We set $\delta_i^p=v_i^p - v_{i-1}^p$ as an axis vector from joint $v_{i-1}^p$ to joint $v_i^p$. Upon this, we define the instantaneous absolute angle as follows \cite{aslanov2012dynamics}:
\begin{equation}
\theta_i^p  =  \overbrace{\theta_{i-1}^p}^{convected} + \underbrace{\widetilde\theta_i^p}_{relative} \\
 = \theta_{i-1}^p + \arccos(\frac {\delta_i^{p+1} \cdot \delta_i^{p}}{ |\delta_i^{p+1}| \cdot  |\delta_i^{p}|}),
\label{eq1}
\end{equation}
where $\theta_{i-1}^p$ is the instantaneous convected rotation angle of joint $v_{i-1}^p$ w.r.t.\ the fixed coordinate system, and 
$\widetilde\theta_i^p$ is the instantaneous relative rotation angle as opposed to joint $v_{i-1}^p$.

\textbf{Instantaneous Angular Velocity.} \quad Further, we obtain the instantaneous angular velocity $\omega_i^p$  of joint $v_i^p$ by differentiating  Equation~(\ref{eq1}) w.r.t.\ $t$:
\begin{equation}
    \omega_i^p = \frac {\mathrm{d} \theta_i^p}{\mathrm{d} t}=\overbrace{\omega_{i-1}^p}^{\mathrm{d} \theta_{i-1}^p/\mathrm{d} t} + \underbrace{\widetilde\omega_i^p}_{\mathrm{d} \widetilde\theta_i^p / \mathrm{d} t}.
\label{eq2}
\end{equation}

\textbf{Instantaneous Angular Acceleration.} \quad To facilitate understanding, we visualize \emph{instantaneous angle} $\theta_i^p$ and \emph{instantaneous angular velocity} $\omega_{i}^p$  in  Figure~\ref{fig2}. Subsequently, we acquire the angular acceleration from the instantaneous angular velocity.
Mathematically, we formulate the angular acceleration of joint $v_i^p$ as $\varepsilon_i^p$.
Naively, we could obtain the acceleration by  differentiating  Equation~(\ref{eq2}) w.r.t.\ $t$, yielding:
\begin{equation}
    \varepsilon_i^p=\frac {\mathrm{d}\omega_i^p }{\mathrm{d} t}= \underbrace{\varepsilon_{i-1}^p}_{\mathrm{d}\omega_{i-1}^p/\mathrm{d} t} + \overbrace{\widetilde\varepsilon_{i}^p + \omega_{i-1}^p \times \omega_i^p}^{\mathrm{d} \widetilde\omega_i^p/\mathrm{d} t},
    \label{eq3}
\end{equation}
where $\varepsilon_i^p$ measures the angular acceleration of joint $v_i^p$.   It is evident to see that the derivation of the above process also applies to computing the angular acceleration of bones.

\subsection{Stream-GCN with Multi-stream Representation Fusion and Channel Attention}

\begin{figure*}[t]
\centering
\includegraphics[width=0.7\textwidth]{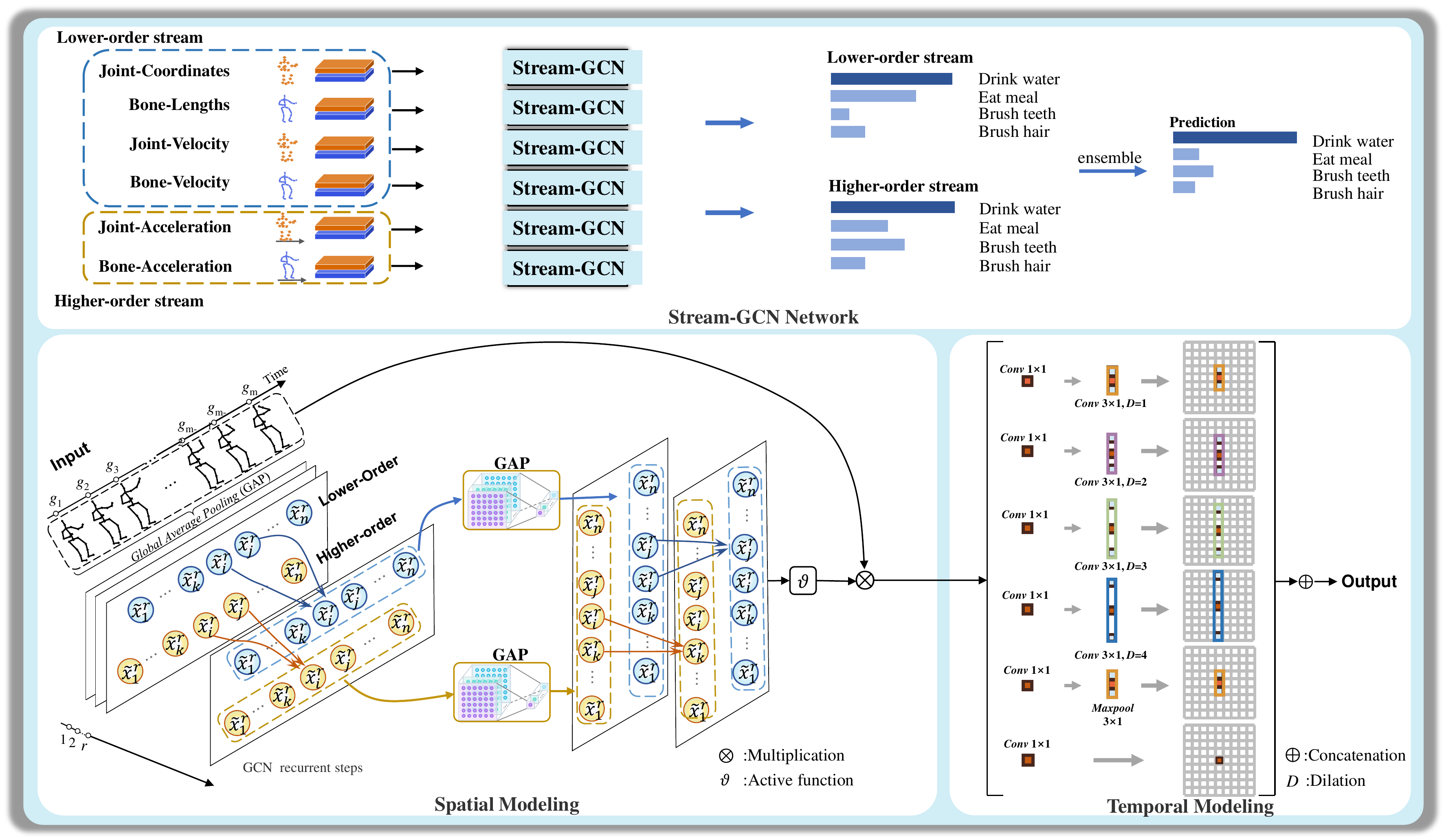} 
\caption{The overall pipeline of our Stream-GCN Network. The goal is to identify the action class by virtue of the motion sequence. For clarity of illustration, we only show one-layer spatial and temporal modeling in this figure. We first absorb input streams from the lower-order and high-order representations. For each stream, the spatial modeling conducts a channel attention module that consists of a set of pooling and convolution operations, yielding the features with channel weights. The temporal modeling adopts multi-scale convolutions, capturing long-range temporal dependencies within a motion sequence. For multi-stream fusion, each stream predicts an action class distribution, which is ensembled to approach the final class distribution.}
\label{fig3}
\end{figure*}

Up to this point, we have presented how to obtain higher-order motion representations. Next, we introduce our Stream-GCN ({GCN} with {{m}}ulti-stream representation fusion and {{c}}hannel attention) network that extracts cues from higher-order and lower-order motion representations to recognize an action. The overall architecture of the network is shown in Figure~\ref{fig3}. The network absorbs six input streams: four lower-order motion representations (joint coordinates, bone lengths, joint velocity, and bone velocity) and two higher-order motion representations (joint and bone angular accelerations). Each input stream is processed by a Stream-GCN network to predict an action class distribution. Their class distributions are then ensembled to give the final class distribution.

\textbf{Stream-GCN Network Architecture.}\quad
Typically, the operation of a GCN layer is given by:
\begin{equation}
X^{r+1} = \sigma(\hat{A}X^rM^r),
\label{eq4}
\end{equation}
where $X^{r} \in \mathbb{R}^{n \times l_r}$ and $X^{r+1} \in \mathbb{R}^{n \times l_r}$ are the features of $r^{th}$ and $r+1^{th}$ layers, respectively.
$n$~is the number of nodes in the graph,
which translates to the number of joints in this task.
$l_r$~is the length of joint features at the $r^{th}$ layer. $\sigma(\cdot)$ is an activation function, \emph{e.g.}, ReLU.
Matrix $M^r \in \mathbb{R}^{l_r \times l_{r+1}}$ is a network parameter (transformation matrix).
Filter matrix $\hat{A}$ is computed based on $\hat{A} = \tilde{D}^{-1/2}\tilde{A}\tilde{D}^{-1/2}$,
where $\tilde{A}=A+I$ and adjacency matrix $A \in \mathbb{R}^{n \times n}$
characterizes the connections between joints.
In the adjacency matrix $A$,
$A_{i,j}=1$ if there exist a bone connects the $i^{th}$ and $j^{th}$ joints, and $A_{i,j}=0$ otherwise.
$\tilde{D} \in \mathbb{R}^{n \times n}$ is
a degree matrix and $\tilde{D}_{i,i} = \sum_j\tilde{A}_{i,j}$.

\textbf{Spatial Modeling.}\quad
Conventional methods such as~\cite{shi2019skeleton,chen2021channel}
directly adopt equal channel weights for all joints and bones
in each frame.
This scheme fails to attend to the distinct roles of different channels in recognizing an action.
Motivated by this, we propose a cross-channel
attention module to adaptively assign weights to different channels.
Formally, in our settings, the input feature is 
$X \in \mathbb{R} ^ {n \times c \times m}$, where
$c$ is the number of channels and $m$ is the number of frames in the given motion sequence. Put differently, the initial feature length (at the $0^{th}$ layer)  $l_0= c \times m$.

Different from Equation (\ref{eq4}), upon the $r^{th}$ layer, we apply one layer graph convolution as below:
\begin{equation}
X^{r+1} = \sigma\big(\hat{A} \tau(X^r) M^r\big),
\label{eq5}
\end{equation}
where $\tau(\cdot)$ is a
cross-channel attention function
that boosts the synergy of different channels. More specifically,  to compute the weights of different channels,  $\tau(\cdot)$ consists of a set of pooling, convolution, and subtraction operations, formulated in Equations (\ref{eq6})--(\ref{eq9}).

Formally, we first apply linear transformation $\mathcal{L}$,  which is a $1\times1$ convolution,
to $X^r$, transforming $X^r$ from  $\mathbb{R} ^ {n \times c_{r} \times m}$ to $\mathbb{R} ^ {n \times c_{r}' \times m}$.
In favor of better aggregating spatial features,
we apply a global pooling operation $\Psi(\cdot)$
in the temporal domain on
features $X^r$ to sketch its spatial features:
\begin{equation}
    \tilde{X}^r =\Psi\big(\mathcal{L}(X^r)\big),
    \label{eq6}
\end{equation}
where $\tilde{X}^r \in \mathbb{R} ^ {n \times c_{r}'}$
is a set of joint spatial  features, namely $\tilde{X}^r = \{\tilde{x}_{1}^r, \tilde{x}_{2}^r, \dots, \tilde{x}_{n}^r\}$ and $\tilde{x}_{i}^r \in \mathbb{R} ^ {c_{r}'}$.

Next, we explore to capture the channel interaction across joint spatial features.
As shown in Figure~\ref{fig3},
given a pair of joints $(v_i, v_j)$
and their corresponding joint spatial features ($\tilde{x}_{i}^r, \tilde{x}_{j}^r$).
We design a correlation modeling function $\mathcal{Z}(\cdot)$
that aggregates channel-wise features
between $v_i$ and $v_j$~\cite{chen2021channel}:
\begin{equation}
    \mathcal{Z}(\tilde{x}_{i}^r, \tilde{x}_{j}^r)=\tilde{x}_{i}^r - \tilde{x}_{j}^r.
    \label{eq7}
\end{equation}

To better exploit the contribution of each channel,
we leverage a global pooling operation $\Phi(\cdot)$ on $ \mathcal{Z}(\tilde{x}_{i}^r, \tilde{x}_{j}^r)$
to obtain a channel feature
$\Phi(\mathcal{Z}(\tilde{x}_{i}^r, \tilde{x}_{j}^r)) \in \mathbb{R}^{c_{r}'}$.
Thereafter, we employ \textit{attention mechanism}~\cite{wang2020ecanet}
on the channel feature to quantify the contributions of
different channels in action recognition:
\begin{equation}
    \vartheta\bigg(\mathcal{F}\Big(\Phi\big(\mathcal{Z}(\tilde{x}_{i}^r, \tilde{x}_{j}^r)\big), k\Big)\bigg),
    \label{eq8}
\end{equation}
where $\vartheta$ is an activation function,
and $\mathcal{F}$ is a $1D$ convolution with kernel size of $k$.
In summary, cross-channel attention function  $\tau(\cdot)$ has the form:
\begin{equation}
    \tau(X^r)=X^r\vartheta\bigg(\mathcal{F}\Big(\Phi\big(\mathcal{Z}(\tilde{x}_{i}^r, \tilde{x}_{j}^r)\big), k\Big)\bigg).
    \label{eq9}
\end{equation}
\textbf{Temporal Modeling.}\quad
To model the action in the temporal domain, we design a multi-scale temporal modeling that follows~\cite{liu2020disentangling}. As shown in Figure~\ref{fig3}, the module consists of six parallel temporal convolutional branches. Each branch starts with a $1 \times 1$ convolution to aggregate features between different channels. 
The first four branches contain a $3 \times 1$ temporal convolution with four different dilations to obtain multi-scale temporal fields. The Maxpool of the fifth branch is applied to emphasize the most significant feature information among consecutive frames. The last $1 \times 1$ convolution is added to the residual-preserving gradient during backpropagation. 
\subsection{Mutual Information Objective}
\label{sec2.3}
We can certainly train the Stream-GCN  with CE loss, as is done in most previous methods ~\cite{ye2020dynamic,cheng2020decoupling}. Given our systematic examination of extracting motion features for action recognition, it would be fruitful to investigate whether introducing a novel supervision method at the feature level would facilitate the task.

Let $Z$ denote the extracted deep features from input $X$. Inspired by ~\cite{liu2022temporal,tian2021farewell}, we try to enforce the correlation between extracted deep features $Z$ and action label $Y$, seeking to  maximize the extraction of task-relevant information and get rid of task-irrelevant nuisances.

\textbf{Mutual Information (MI)}\quad
MI measures the amount of information shared between two random variables. Formally, MI between $Z$ and $Y$ quantifies the statistical dependency of encoding variables $Z$ and action label $Y$:
\begin{equation}
    \mathcal{I}(Y; Z)= \mathbb{E}_{p(Y,Z)}[log \frac{p(Y,Z)} {p(Y)p(Z)}],
    \label{eq10}
\end{equation}
where $p(Y,Z)$ is the probability distribution between $Z$ and $Y$, while $p(Z)$ and $p(Y)$ are their marginals   \cite{liu2022temporal}.

\textbf{Mutual Information Loss}\quad
In order to ensure that the maximum amount of information about $Y$ is extracted to the deep feature $Z$, while reducing the information that is not relevant to $Y$ in $Z$, our primary objective can be formulated as:
\begin{equation}
    {\rm IB} (Y,Z)=\max \mathcal{I}(Y; Z|X).
    \label{eq11}
\end{equation}

\begin{figure}[t]
\centering
\includegraphics[width=0.25\textwidth]{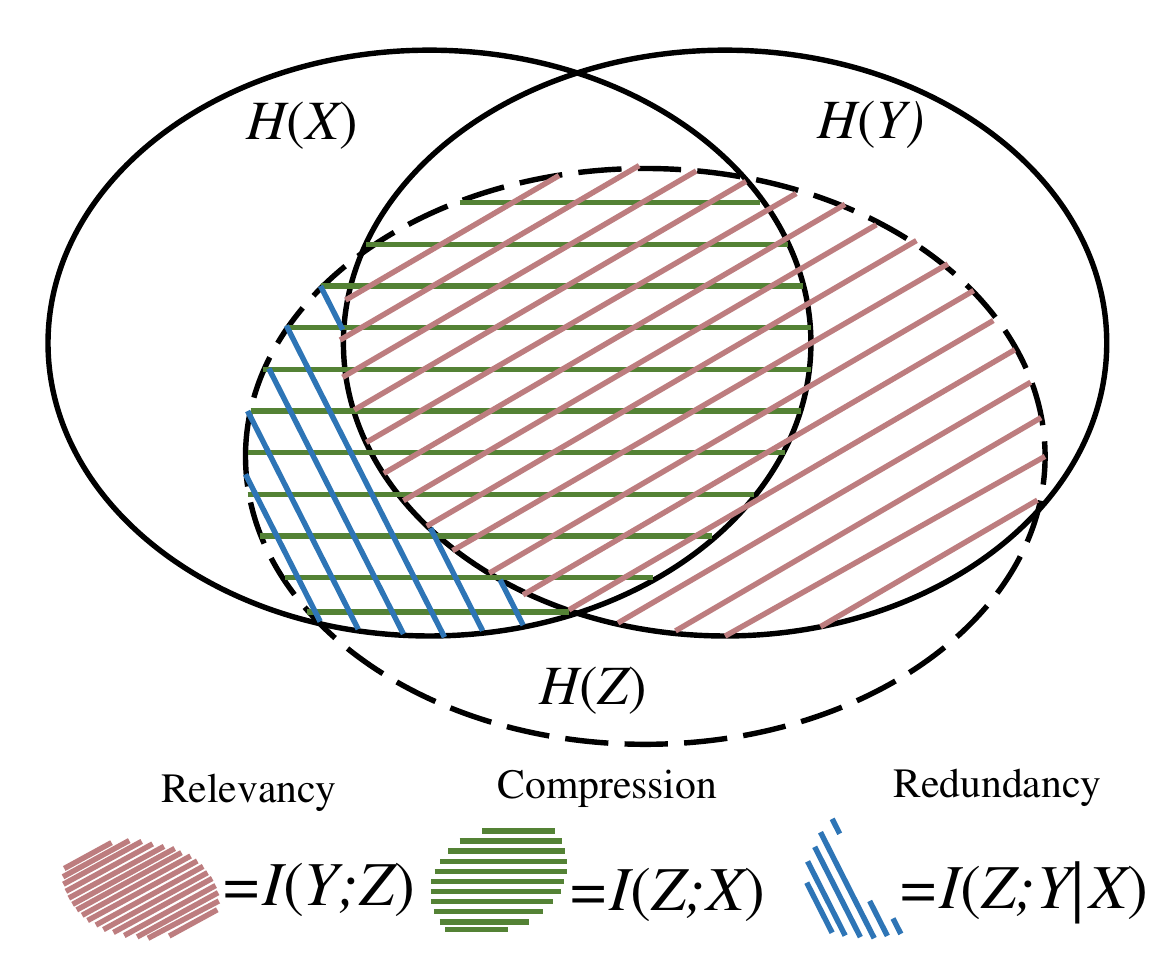} 
\caption{Our mutual information maximizes $I(Y;Z)$, compresses $I(Z;X)$, and preserves $I(Z;Y|X)$. $H(\cdot)$ denotes entropy. The visualization is inspired by~\protect\cite{yeung1991new}.} \vspace{-1em}
\label{fig4}
\end{figure}

Due to the notorious difficulty of the conditional MI  computations especially in neural networks ~\cite{huang2020devil,tian2021farewell}, we perform  a simplification.  We factorize Equation (\ref{eq11}) as follows:
\begin{equation}
    {\rm IB} (Y,Z) =  \overbrace{\mathcal{I}(Y; Z)}^{relevancy}- \underbrace{[\mathcal{I}(Z; X)]}_{compression}+ \overbrace{[\mathcal{I}(Z;Y|X)]}^{redundancy},
\label{eq12}
\end{equation}
The first term forces features to be informative about the label $Y$. The second term compresses $Z$ to be concise. The third term constraints $Z$ to include task-relevant information in $X$. These MI terms are depicted in Figure~\ref{fig4} and can be estimated by existing MI estimators~\cite{belghazi2018mutual,cheng2020club,tian2021farewell}. In our experiments, we employ the Variational Information Bottleneck (VIB)~\cite{alemi2016deep} and Variational Self-Distillation (VSD)~\cite{tian2021farewell} to estimate  each MI term.

\section{Experiments}
In this section, we conduct  extensive experiments to  empirically evaluate our method
on three benchmark action recognition datasets. Broadly, we intend to answer the following research questions:
\begin{itemize}
\item \textbf{RQ1:} How does the proposed method compare against the state-of-the-art approaches for skeleton-based action recognition?
\item \textbf{RQ2:} How much do different components of Stream-GCN contribute to its performance?
\item \textbf{RQ3:} What interesting insights and findings can we obtain from the empirical results?
\end{itemize}
Next, we first present the experimental settings, followed by answering the above research questions one by one.

\subsection{Experimental Settings}
\textbf{Datasets.}\quad
We adopt three widely used action recognition datasets,
namely NTU-RGB+D, NTU-RGB+D 120, and Northwestern-UCLA,
to evaluate the proposed method. 

\emph{NTU-RGB+D.} \quad
NTU-RGB+D~\cite{shahroudy2016ntu} is tailored for the skeleton-based action recognition task.
It contains 56,880 video samples from 60 action classes performed by 40 volunteers. Each sample contains one action and is guaranteed to have a maximum of two subjects, which is captured by three Microsoft Kinect v2 cameras.
This dataset provides two sub-benchmarks:
(1)~ Cross-Subject (X-Sub): data for 20 subjects is used as the training data, while the rest is used as test data.
(2)~ Cross-View (X-View) divides the training and test sets
according to different camera views. 

\emph{NTU-RGB+D 120.}\quad
NTU-RGB+D 120~\cite{liu2019ntu} is the current largest human action recognition dataset,
which extends NTU-RGB+D with an additional 60 action classes and 57,600 video samples.
It contains a total of 114,480 samples from 120 action classes, which are executed by 106 volunteers and captured by three Kinect cameras.
Within the dataset, two benchmarks are maintained:
(1) Cross-Subject (X-Sub), which categorizes 53 subjects
into the training class and the other 53 subjects into the test class.
(2) Cross-Setup (X-Set), which arranges data items with even IDs into the training group
and odd IDs into the test group.

\emph{Northwestern-UCLA.} \quad 
The Northwestern-UCLA dataset contains 1,494 video samples in 10 classes~\cite{wang2014cross}.
These videos were collected by filming ten actors with three Kinect cameras.
We follow the evaluation protocol mentioned in~\cite{wang2014cross}, where videos collected by the first two cameras serve as the training samples and the rest serve as test samples.

\textbf{Implementation Details.} \quad
We conduct experiments on a computer
equipped with an Intel Xeon E5 CPU at 2.1GHz,
three NVIDIA GeForce GTX 1080 Ti GPUs, and the RAM of 64GB.
We leverage PyTorch 1.1 to implement our model.
We apply pre-processing method of~\cite{zhang2020semantics} to
the data in NTU-RGB+D and NTU-RGB+D 120.
Moreover, we initialize the data in Northwestern-UCLA
according to the approach of~\cite{cheng2020skeleton}. 

We apply stochastic gradient descent (SGD)
with 0.9 Nesterov momentum to train the Stream-GCN model.
For NTU-RGB+D and NTU-RGB+D 120 datasets, the number of training epochs
is set to 65 with the first 5 epochs being warm-up epochs, which help stabilize the training process.
For NTU-RGB+D and NTU-RGB+D 120 datasets, the initial learning rate is set to 0.1 and decays by 0.1 every 35 epochs, the batch size is selected as 64.  For the Northwestern-UCLA dataset,  the initial learning rate is set to 0.01 and decays by 0.0001 every 50 epochs, the batch size is set to 16.

\subsection{Comparison with Existing Methods (RQ1)}

We first empirically compare the proposed model with the state-of-the-art methods.
The experimental results are summarized in Tables~\ref{table1}--\ref{table3}.
\begin{table}[t]
  \centering
  \resizebox{.98\columnwidth}{!}{
  \begin{tabular}{lcc}
    \toprule
    Methods & X-Sub(\%) & X-Set(\%)\\
    \midrule
    ST-LSTM~\cite{liu2016spatio} & 55.7 & 57.9 \\
    GCA-LSTM~\cite{liu2017skeleton} & 61.2 & 63.3 \\
    \midrule
    2s-AGCN~\cite{shi2019two} & 82.9 & 84.9 \\
    4s Shift-GCN~\cite{cheng2020skeleton} & 85.9 & 87.6 \\
    DC-GCN+ADG~\cite{cheng2020decoupling} & 86.5 & 88.1 \\
    MS-GCN~\cite{liu2020disentangling} & 86.9 & 88.4 \\
    Dynamic GCN~\cite{ye2020dynamic} & 87.3 & 88.6 \\
    MST-GCN~\cite{feng2021multi} & 87.5 & 88.8 \\
    Ta-CNN~\cite{xu2022topology} & 85.4 & 86.8 \\
    EfficientGCN-B4~\cite{song2022constructing} & 88.3 & 89.1 \\ 
    \midrule
    Stream-GCN(Joint stream)  & 85.8 & 86.7 \\
    Stream-GCN(Bone stream)  & 86.5 & 88.3 \\
    Stream-GCN(4 streams)  & 89.2 & 90.4 \\
    Stream-GCN(6 streams)  & \textbf{89.7} & \textbf{91.0} \\
    \bottomrule
    \end{tabular}}
    \caption{Comparisons of the Top-1 accuracy(\%) with the state-of-the-art methods on the NTU RGB+D 120 dataset.}\vspace{-0.5em}
    \label{table1}
\end{table}
Table~\ref{table1} and Table~\ref{table2} present results on NTU RGB+D 120 and NTU RGB+D datasets, while Table~\ref{table3} illustrates results on the Northwestern-UCLA dataset. 
From the tables, we have the following observations. \textbf{(a)} The proposed method consistently outperforms state-of-the-art approaches on all three benchmark datasets. For instance, on the NTU120 RGB+D dataset, state-of-the-art method~\cite{song2022constructing} achieves a 88.3\% action recognition accuracy. In contrast, Stream-GCN obtains a 89.7\% accuracy. 
\begin{table}[t]
  \centering
  \resizebox{.98\columnwidth}{!}{
  \begin{tabular}{lcc}
    \toprule
    Methods & X-Sub(\%) & X-View(\%) \\
    \cmidrule(r){1-3}
    IndRNN~\cite{liu2016spatio} & 81.8 & 88.0 \\
    HCN~\cite{liu2017skeleton} & 86.5 & 91.1 \\
    \midrule
    2s-AGCN~\cite{shi2019two} & 88.5 & 95.1 \\
    SGN~\cite{zhang2020semantics} & 89.0 & 94.5 \\
    AGC-LSTM~\cite{si2019attention} & 89.2 & 95.0 \\
    DGNN~\cite{shi2019skeleton} & 89.9 & 96.1 \\
    4s Shift-GCN~\cite{cheng2020skeleton} & 90.7 & 96.5 \\
    DC-GCN+ADG~\cite{cheng2020decoupling} & 90.8 & 96.6 \\
    Dynamic GCN~\cite{ye2020dynamic} & 91.5 & 96.0 \\
    MS-GCN~\cite{liu2020disentangling} & 91.5 & 96.2 \\
    MST-GCN~\cite{feng2021multi} & 91.5 & 96.6 \\
    Ta-CNN~\cite{xu2022topology} & 90.4 & 94.8 \\
    EfficientGCN-B4~\cite{song2022constructing} & 91.7 & 95.7 \\
    \midrule
    Stream-GCN  & \textbf{92.9} & \textbf{96.9} \\
    \bottomrule
    \end{tabular}}
    \caption{Comparisons of the Top-1 accuracy(\%) with the state-of-the-art methods on the NTU RGB+D dataset.}\vspace{-0.2em}
  \label{table2}
\end{table}
\begin{table}[t]
  \centering
  \begin{tabular}{lcc}
    \toprule
    Methods & Top-1(\%) \\
    \cmidrule(r){1-2}
    Lie Group~\cite{veeriah2015differential} & 74.2 \\
    Ensemble TS-LSTM~\cite{lee2017ensemble} & 89.2 \\
    \midrule
    4s Shift-GCN~\cite{cheng2020skeleton} & 94.6 \\
    DC-GCN+ADG~\cite{cheng2020decoupling} & 95.3 \\
    Ta-CNN~\cite{xu2022topology} & 96.1 \\
    \midrule
    Stream-GCN  & \textbf{96.8} \\
    \bottomrule
    \end{tabular}
    \caption{Comparisons of the Top-1 accuracy(\%) with the state-of-the-art methods on the Northwestern-UCLA dataset.}\vspace{-0.5em}
    \label{table3}
\end{table}
\textbf{(b)} Generally, GCN based approaches (such as ~\cite{shi2019skeleton,shi2019two,liu2020disentangling}) perform better than RNN based methods (such as ~\cite{liu2016spatio,liu2017skeleton,liu2017skeleton}). The empirical evidence suggests that GCN models are more suitable for action recognition tasks as they may conveniently model the graph structure of the human body.  \textbf{(c)} Another finding is that multi-stream models usually outperform single-stream ones. For example, on the NTU RGB+D 120 dataset, multi-stream models~\cite{cheng2020skeleton,song2022constructing,ye2020dynamic} achieve much better results than single stream methods~\cite{zhang2020semantics,si2019attention}. This reveals the importance of incorporating different representations to approach more precise action recognition.


\subsection{Ablation Study (RQ2)}

We perform ablation experiments to examine the influence of each component in our method (\emph{i.e.}, Motion Representation, Cross-channel attention Module, and MI Loss). All the ablation studies are conducted on the NTU-RGB+D 120 cross-subject benchmark dataset.  We also investigate the impact of modifying the parameter settings of the cross-channel attention module. 

\textbf{Motion Representation.}\quad To verify the impact of the proposed acceleration representations in modeling higher-order motion features. We tried removing the joint angular acceleration stream and bone angular acceleration stream respectively. Specifically, we \emph{first} tried removing all acceleration streams. \emph{Then}, we studied removing only the acceleration stream of the joint. \emph{Finally}, we also tried removing only the acceleration stream of bone. The quantitative results are demonstrated in Table~\ref{table4}, empirical results reveal that using the high-order representation indeed boosts the action recognition accuracy.

\begin{table}
\centering
\resizebox{.85\columnwidth}{!}{
\begin{tabular}{cccc} 
  \toprule
\multirow{2}{*}{Ensemble Methods} & \multicolumn{2}{c}{Acceleration Streams}  & \multirow{2}{*}{Acc(\%)}  \\ 
\cmidrule{2-3}
                   &    Joint    &     Bone     &                     \\ 
\midrule
4 Streams          &             &              &   89.2           \\
5 Streams          &             &  \Checkmark  &   89.4             \\
5 Streams          &  \Checkmark &              &   89.3           \\
6 Streams          &  \Checkmark &  \Checkmark  &   \textbf{89.7}      \\
\bottomrule
\end{tabular}}
  \caption{Study on removing acceleration streams.}\vspace{-0.3em}
  \label{table4}
\end{table}

\begin{table}[t]
  \centering
  \resizebox{.9\columnwidth}{!}{
  \begin{tabular}{cccc}
    \toprule
     Attention Module  & Mutual Information & Params & Acc(\%) \\
    \midrule
      &  & 1.21M & 88.9 \\
	
      & \Checkmark & 1.23M & 89.4 \\
	\Checkmark &   & 1.64M & 89.3\\
    \Checkmark & \Checkmark & 1.65M & \textbf{89.7}\\
    \bottomrule
  \end{tabular}}
  \caption{The impact of each component.}\vspace{-0.5em}
  \label{table5}
\end{table}

\begin{table}[t]
  \centering
  \resizebox{.96\columnwidth}{!}{
  \begin{tabular}{lccccccc}
    \toprule
    \multirow{2}{*}{Methods}  & \multirow{2}{*}{$\vartheta$} & \multicolumn{5}{c}{Kernel size} \\
    \cmidrule{3-7}
     &   &  $K=1$ & $K=3$ & $K=5$ & $K=7$ & $K=9$ \\
    \midrule
    Stream-GCN & Sig($\cdot$) & \textbf{85.8} & 84.0 & 84.3 & 84.0 & 79.9\\
    Stream-GCN + Mask & Sig($\cdot$) & 84.4 & 84.1 & 84.4 & 84.3 & 83.9\\
    \midrule
    Stream-GCN & ReLU($\cdot$) & 84.1 & 84.1 & 84.2 & 84.0 & 84.2 \\
    Stream-GCN + Mask & ReLU($\cdot$) & 84.0 & 83.9 & 84.3 & 84.0 & 84.3 \\
    \midrule
    Stream-GCN & Tanh($\cdot$) & 83.7& 84.3 & 84.0 & 84.1 & 84.0 \\
    Stream-GCN + Mask & Tanh($\cdot$) & 83.9 & 84.4 & 84.2 & 83.9 & 83.9 \\
    \bottomrule
    \end{tabular}}
    \caption{Accuracy (\%) with different parameter settings on the cross-channel attention module.}\vspace{-0.3em}
  \label{table6}
\end{table}

\textbf{Cross-channel attention Module.}\quad Next, we are interested in the effect of removing the cross-channel attention module.
The empirical results are demonstrated in Table~\ref{table5}.  Remarkably, the result in the second line of Table~\ref{table5} shows that when the cross-channel attention module is removed, the accuracy is decreased (89.7\% vs 89.4\%).

\textbf{Mutual Information Loss.}\quad We validate the effect of the MI Loss in Section~\ref{sec2.3}. As illustrated in the third and the fourth lines of Table~\ref{table5}, the 0.4\%  accuracy improvement provides empirical evidence that our proposed MI loss is effective.  This reveals that the additional feature-level supervision facilitates the extraction of task-specific information.

\textbf{Parameter Settings.}\quad By default, we use cross-channel attention to assign different weights to different channels in recognizing an action. We explore the hyper-parameter settings in the cross-channel attention module with joint stream (\emph{i.e.}, joint coordinates). The key hyper-parameters include the activation function $\vartheta$, the kernel size of the $1D$ convolution, and the mask on the adjacency matrix. As shown in Table~\ref{table6}, we observe that Stream-GCN obtains the best performance when $\vartheta$ is a Sigmoid function, the kernel size $K=1$, and the adjacency matrix is without a mask.

\subsection{Discussions (RQ3)}

\begin{figure}[ht]
\centering
\includegraphics[width=0.40\textwidth]{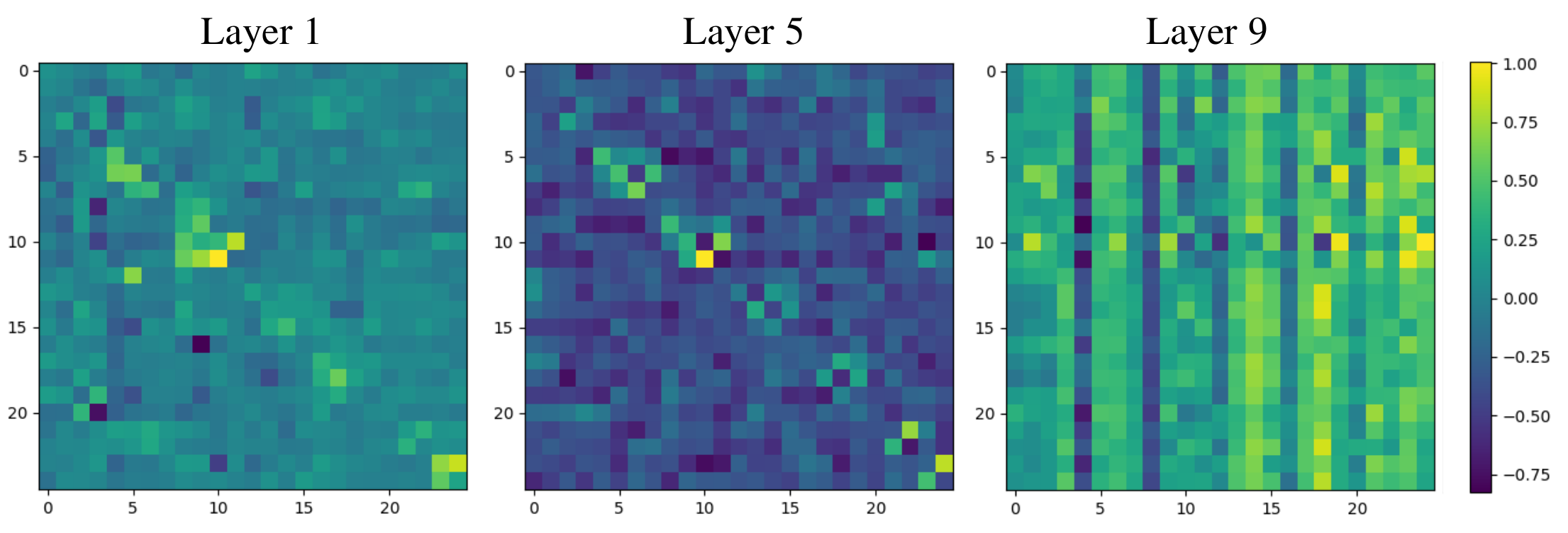} 
\caption{Examples of the learned attention maps at different layers for the drinking water action. The numbers denote different joints, e.g., \emph{``number \rm{4}''} denotes \emph{``head"} and \emph{``number \rm{8}''} denotes \emph{``left hand''}. The brighter area indicates that the weight of the correlation matrix is larger there, which means the correlation strength between the joints is stronger.}\vspace{-0.3em}
\label{fig5}
\end{figure}

\textbf{Visualize Results.} \quad We visualize the learned attention maps of different layers (\emph{layer \rm{1}}, \emph{layer \rm{5}}, and \emph{layer \rm{9}}) for the ``drinking water" action  in Figure~\ref{fig5}. From the figure, we see that different layers of the learned attention map contain distinct  semantics.
Specifically, in the lower layers, it seems that the model focuses on the relations between hands, shoulders, and wrists. This is intuitive since the model will initially focus on the body parts (joints) where the movement takes place. In the intermediate layers, the arm is more concerned by the model, with attention to the other parts gradually diminishing. This demonstrates that the model is analyzing what exactly the arm is doing. In the higher layers, the information is highly aggregated. Meanwhile, the model will combine other joints that are helpful to identify the motion.  As an example, for recognizing the action of ``drinking water", the correlation between the hands and the head is detected and leveraged by the model.

Experiments on three different benchmark datasets suggest that incorporating  higher-order motion sequences achieves higher accuracy. 
Interestingly, a point that attracts our attention is:  we empirically observe that as the number of action classes goes higher, although incorporating  high-order representation still improves the accuracy,  the advantage of incorporating high-order representation decreases. We plan to dive deeper into this phenomenon and come up with new solutions.

\section{Related Work}

Earlier skeleton-based action recognition methods model the motion sequences with convolutional neural networks or recurrent neural networks~\cite{si2018skeleton,liu2017skeleton}. As these conventional methods have inherent difficulties in capturing the connections between joints, they tend to suffer from unsatisfactory results. 
Recent efforts have embraced graph convolutional networks (GCNs) to model the human motion sequences as spatio-temporal graphs~\cite{bian2021structural,chen2021multi}.
\cite{yan2018spatial} is a pioneering work in this line of efforts, proposing a ST-GCN model to characterize the skeleton motion as a GCN. \cite{shi2019two} proposes an adaptive graph convolution network based on ST-GCN, incorporating skeletal bone features as an additional input stream and adopting self-attention to learn the graph parameters.  \cite{shi2019skeleton} introduces frame-wise bone and joint velocities as  additional input streams for action recognition.
Existing methods tend to have difficulties in discriminating between similar actions. Meanwhile, to our best knowledge, the higher-order motion features and mutual information are rarely explored in skeleton-based action recognition. This inspires us to seek novel methods to facilitate the model recognizing the action class precisely. 

\section{Conclusion}
In this paper, we have proposed a motion representation derived from rigid body kinematics. The new representation captures the higher-order motion features, complementing conventional motion representations. We also present a Stream-GCN network equipped with multiple input streams and channel-wise attention, where different streams are ensembled while attention weights capitalize on those important channels. Finally, we introduce a mutual information objective for theoretic supervision on  extracted features. Extensive experiments show that our method consistently surpasses state-of-the-art methods on three different benchmark datasets. Our code is released to facilitate researchers.

\bibliographystyle{named}
\bibliography{ijcai23}

\end{document}